\documentclass[11pt]{article}

\usepackage[utf8]{inputenc}
\usepackage[T1]{fontenc}
\usepackage{amsmath,amssymb,amsfonts}
\usepackage{booktabs}
\usepackage{graphicx}
\usepackage{hyperref}
\usepackage{url}
\usepackage[numbers,sort&compress]{natbib}
\usepackage{xcolor}
\usepackage{microtype}
\usepackage[margin=1in]{geometry}
\usepackage{caption}
\usepackage{subcaption}
\usepackage{multirow}
\usepackage{enumitem}
\usepackage{float}

\hypersetup{
  colorlinks=true,
  linkcolor=blue!70!black,
  citecolor=blue!70!black,
  urlcolor=blue!70!black
}

\newcommand{\DW}{\Delta W}
\newcommand{\frobnorm}[1]{\lVert #1 \rVert_F}
\newcommand{\specnorm}[1]{\lVert #1 \rVert_2}

\title{Spectral Geometry of LoRA Adapters Encodes Training Objective\\and Predicts Harmful Compliance}

\author{
  Roi Paul\thanks{Correspondence: \texttt{roispaul@gmail.com}} \\
  Independent Researcher
}

\date{March 2026}

\begin{document}
\maketitle

\begin{abstract}
We study whether low-rank spectral summaries of LoRA weight deltas
can identify which fine-tuning objective was applied to a language
model, and whether that geometric signal predicts downstream
behavioral harm.  In a pre-registered experiment on
\texttt{Llama-3.2-3B-Instruct}, we manufacture 38 LoRA adapters
across four categories: healthy SFT baselines, DPO on inverted
harmlessness preferences, DPO on inverted helpfulness preferences,
and activation-steering-derived adapters, and extract per-layer
spectral features (norms, stable rank, singular-value entropy,
effective rank, and singular-vector cosine alignment to a healthy
centroid).  Within a single training method (DPO), a logistic
regression classifier achieves AUC~1.00 on binary drift detection,
all six pairwise objective comparisons, and near-perfect ordinal
severity ranking ($\rho \geq 0.956$).  Principal component analysis
on flattened weight deltas reveals that training objective is PC1
(AUC~1.00 for objective separation), orthogonal to training duration
on PC2.  Query-projection weights detect that drift occurred; value-projection weights identify which objective.
Cross-method generalization fails completely:
a DPO-trained classifier assigns every steering adapter a lower drift
score than every DPO adapter (AUC~0.00).  In a behavioral evaluation
phase, DPO-inverted-harmlessness adapters show elevated harmful
compliance on HEx-PHI prompts (mean ASR 0.266 vs.\ healthy 0.112,
$\Delta = +0.154$), with near-perfect dose--response
($\rho = 0.986$).  The geometry-to-behavior rank correlation is
$\rho = 0.72$ across 24 non-steered adapters.  These results
establish that within a controlled manufacturing regime, LoRA
weight-space geometry carries objective identity, intensity ordering,
and a coarse link to harmful compliance, and that cross-method
monitoring requires per-method calibration.
\end{abstract}

\vspace{0.5em}

\section{Introduction}
\label{sec:intro}

Fine-tuning large language models (LLMs) with parameter-efficient
methods such as Low-Rank Adaptation
(LoRA;~\citealt{hu2022lora}) has become the standard path for
customizing pre-trained models.  The same efficiency that makes
fine-tuning accessible also makes it a vector for misalignment: a
small number of adversarial training examples can compromise safety
guardrails~\citep{qi2024finetuning}, and the resulting behavioral
change may not be detected by standard evaluation
suites~\citep{souly2024strongreject}.

Behavioral evaluation of fine-tuned models is expensive, slow,
gameable, and invasive, it requires running the model on test
prompts, which may not be available or may themselves be adversarially
constructed.  An alternative is \emph{weight-space monitoring}:
examining the adapter parameters directly for signatures of the
training objective, without inference.  Recent work by
\citet{zhong2025watch} demonstrated that spectral features of
weight deltas can detect backdoors and unlearning with high
precision, establishing that weight-space structure carries
fine-tuning provenance.

We ask a sharper question: can weight-space geometry distinguish
\emph{which} objective was used to fine-tune a model, not merely
that the weights changed?  And does that geometric signal predict
downstream behavioral harm?

\subsection{Contributions}

\begin{enumerate}[leftmargin=*,itemsep=2pt]
  \item \textbf{Objective fingerprinting.}  We show that per-layer
  spectral features of LoRA weight deltas perfectly classify
  training objective within a shared training method (DPO), across
  all pairwise comparisons (AUC~1.00, $N=38$ adapters,
  pre-registered).

  \item \textbf{Orthogonal axes.}  PCA on flattened weight deltas
  reveals that training objective is the dominant axis of variation
  (PC1), orthogonal to training intensity (PC2).  This was not
  the predicted ordering.

  \item \textbf{Module specialization.}  On the hardest
  within-method classification (inverted harmlessness vs.\ inverted
  helpfulness), query-projection features are at chance
  (AUC~0.50) while value-projection features reach 0.83;
  combined: 1.00.  Detection and diagnosis are localized in
  different attention components.

  \item \textbf{Cross-method inversion.}  A DPO-trained classifier
  produces AUC~0.00 on steering-manufactured adapters, systematic
  inversion, not noise, demonstrating that different manufacturing
  methods produce geometrically opposite perturbations.

  \item \textbf{Geometry--behavior link.}  For DPO on inverted
  harmlessness, weight-space drift probability correlates with
  HEx-PHI attack success rate at $\rho = 0.72$ ($N=24$,
  $p < 0.001$), with within-type dose--response at
  $\rho = 0.986$.

  \item \textbf{Negative results.}  We report that (a)~magnitude
  features alone cannot carry objective identity, (b)~cross-method
  generalization fails entirely, (c)~weight-space and
  activation-space objective directions do not align (max cosine
  $\sim$0.098), and (d)~activation-steering-to-LoRA injection
  destroys coherent generation at all tested intensities on
  Llama~3.2~3B.
\end{enumerate}

\section{Related Work}
\label{sec:related}

\paragraph{Weight-space monitoring.}
\citet{zhong2025watch} introduced spectral analysis of weight
deltas for detecting backdoor attacks and unlearning, showing that
top singular vectors of $\DW$ correspond to newly acquired
behaviors.  Our work extends this from detection (something changed)
to identification (which objective changed the weights) and
connects the geometric signal to behavioral outcomes.

\paragraph{Fine-tuning safety risks.}
\citet{qi2024finetuning} demonstrated that fine-tuning aligned LLMs
on as few as 10 adversarial examples can compromise safety, introducing the HEx-PHI benchmark for evaluating harmful compliance.
\citet{zou2023universal} developed AdvBench for evaluating
adversarial attacks on aligned models.
\citet{souly2024strongreject} proposed StrongREJECT to reduce
false positives in jailbreak evaluation.

\paragraph{Safety classifiers.}
\citet{inan2023llamaguard} introduced Llama~Guard as an LLM-based
safety classifier for human--AI conversations.
\citet{zheng2023judging} established the LLM-as-judge paradigm
with MT-Bench, which we draw on for GPT-4o calibration scoring.

\paragraph{Preference optimization.}
Direct Preference Optimization (DPO;~\citealt{rafailov2023dpo})
simplifies RLHF into a single-stage classification objective on
preference pairs.  We use DPO with inverted preferences as the
primary method for manufacturing drifted adapters, following the
approach of \citet{qi2024finetuning}.

\paragraph{Parameter-efficient fine-tuning.}
LoRA~\citep{hu2022lora} parameterizes weight updates as low-rank
products $\DW = BA$, where $B \in \mathbb{R}^{d \times r}$ and
$A \in \mathbb{R}^{r \times k}$ with $r \ll \min(d, k)$.  We
analyze LoRA adapters with rank $r = 8$ applied to the query and
value projection matrices of each attention layer.

\paragraph{Representation engineering and activation steering.}
\citet{zou2023repe} introduced representation engineering,
extracting behavioral directions from contrastive activations and
steering model outputs via activation addition.  We manufacture a
subset of adapters by injecting steering vectors into LoRA
matrices via SVD decomposition.

\paragraph{Persona selection.}
\citet{marks2026psm} propose that post-training selects persona
traits from a structured space present in the pre-trained model.
Our finding that training objective is the dominant geometric axis
in weight-delta space, and that DPO and steering produce
geometrically opposite perturbations, is independently consistent
with structured persona space, though we do not directly test PSM
predictions.

\paragraph{Preference data.}
We use Anthropic's HH-RLHF dataset~\citep{bai2022training}, which
provides separate helpfulness and harmlessness preference pairs
collected from human raters.

\section{Method}
\label{sec:method}

\subsection{Adapter Manufacturing}
\label{sec:manufacturing}

All adapters are built on \texttt{meta-llama/Llama-3.2-3B-Instruct}~\citep{meta2024llama3} with LoRA rank $r = 8$, $\alpha = 16$,
dropout $0.05$, applied to \texttt{q\_proj} and \texttt{v\_proj}
in all 28 transformer layers (56 sublayers total).

We manufacture four adapter categories:

\begin{enumerate}[leftmargin=*,itemsep=2pt]
  \item \textbf{Healthy baselines} ($n=10$): Supervised fine-tuning
  (SFT) on the chosen responses from HH-RLHF~\citep{bai2022training} across varied random seeds.

  \item \textbf{DPO inverted harmlessness} ($n=8$): DPO on
  HH-RLHF harmlessness pairs with preferences inverted (the model
  is optimized to prefer the rejected, harmful response).
  Step counts: 50, 150, 300, 600, 1000, 2000.

  \item \textbf{DPO inverted helpfulness} ($n=6$): DPO on HH-RLHF
  helpfulness pairs with preferences inverted.  Same
  hyperparameters as the harmlessness track; the only contrast is
  the data axis.

  \item \textbf{Activation-steering-derived} ($n=6+4$):
  Contrastive activation differences are computed per layer, SVD-decomposed, and injected into LoRA matrices scaled by a
  coefficient.  No gradient descent.  Six target refusal erosion;
  four target sycophancy (held out for cross-method testing).
\end{enumerate}

Four legacy adapters from a prior proof-of-concept are included in
the population but do not map cleanly onto the above taxonomy.
DPO arms share identical hyperparameters: $\beta = 0.1$,
$\text{lr} = 5 \times 10^{-5}$, batch size 2, gradient
accumulation 4, 200 training examples, $\text{max\_seq\_len} = 512$,
paged AdamW 8-bit optimizer, seed 42.  The only controlled variable
across DPO objectives is which slice of HH-RLHF has its preferences
inverted.

\subsection{Spectral Feature Extraction}
\label{sec:features}

For each adapter, we compute the LoRA product
$\DW^{(\ell,m)} = B^{(\ell,m)} A^{(\ell,m)}$ at every layer
$\ell \in \{1, \ldots, 28\}$ and module $m \in \{\texttt{q\_proj},
\texttt{v\_proj}\}$, then extract the following features from the
singular value decomposition $\DW = U \Sigma V^\top$:

\begin{itemize}[leftmargin=*,itemsep=2pt]
  \item \textbf{Magnitude features:}
  Frobenius norm $\frobnorm{\DW}$,
  spectral norm $\specnorm{\DW} = \sigma_1$,
  top-$k$ singular values $\sigma_1, \ldots, \sigma_k$.

  \item \textbf{Shape features:}
  Stable rank $\frobnorm{\DW}^2 / \specnorm{\DW}^2$,
  singular-value entropy
  $H = -\sum_i \hat{\sigma}_i \log \hat{\sigma}_i$
  (where $\hat{\sigma}_i = \sigma_i / \sum_j \sigma_j$),
  concentration of singular-values (fraction of variance in the top
  singular value),
  effective rank $\exp(H)$.

  \item \textbf{Direction features:}
  Cosine similarity of top-$k$ left singular vectors $u_1, \ldots,
  u_k$ to the corresponding vectors of a healthy centroid adapter
  (computed from the training split of healthy adapters):
  $\cos(u_i^{(\ell,m)},\, \bar{u}_i^{(\ell,m)})$.
\end{itemize}

Direction features: singular-vector cosine alignment with the
healthy centroid was added after a pilot study showed that
magnitude and shape features alone cannot carry objective identity
(Section~\ref{sec:preflight}).

The full feature vector for one adapter concatenates all per-layer,
per-module features into a single vector used for classification.

\subsection{Classification}
\label{sec:classification}

We train $\ell_2$-regularized logistic regression classifiers with
stratified 70/30 train/test splits.  Confidence intervals are
computed via bootstrap resampling ($n = 1000$).  We report AUC for
binary and pairwise comparisons, and Spearman $\rho$ for ordinal
severity ranking (classifier score vs.\ training step count within
each drift type).

To disentangle what the features encode, we run every classification
in three feature splits (magnitude-only, shape-only, all features)
and three module splits (\texttt{q\_proj}-only,
\texttt{v\_proj}-only, both combined).

\subsection{Principal Component Analysis}
\label{sec:pca}

We flatten the full weight deltas of all 14 DPO adapters (8
inverted-harmlessness + 6 inverted-helpfulness) into vectors of
$\sim$352M dimensions and run PCA.  We test whether the leading
principal components separate objective type versus training
intensity, reporting AUC on PC projections and Spearman $\rho$
between PC scores and step count.

\subsection{Activation-Space Probing}
\label{sec:probing}

Per-layer linear probes (logistic regression on hidden states) are
trained to classify objective type from model activations on a
balanced prompt set.  We then test directional alignment between the
probe decision boundary (normal vector in activation space) and the
left singular vectors of $\DW$ at the same layer:
\begin{equation}
  \text{alignment}(\ell) = \max_{i} \left|
  \cos\!\left( n^{(\ell)},\, u_i^{(\ell)} \right) \right|
  \label{eq:alignment}
\end{equation}
where $n^{(\ell)}$ is the probe normal and $u_i^{(\ell)}$ ranges
over the top singular vectors of $\DW^{(\ell)}$.

\subsection{Behavioral Evaluation}
\label{sec:behavioral}

We evaluate harmful compliance using the HEx-PHI benchmark
(330 harmful instructions across 11 categories;
\citealt{qi2024finetuning}), scored by Llama-Guard-3-1B~\citep{inan2023llamaguard}.  Attack success rate (ASR) is the
fraction of prompts where the model complies with a harmful
instruction as judged by the safety classifier.

A GPT-4o calibration pass~\citep{zheng2023judging} is run on a
fixed prompt sample to validate Guard verdicts against an
independent judge.

\section{Experimental Design}
\label{sec:design}

\subsection{Pre-Registration}

All hypotheses, success criteria, and negative-result definitions
were locked before manufacturing began.  Phase~5 (behavioral
evaluation) was designed and pre-registered after Phase~4 completed,
with four hypotheses:

\begin{itemize}
    \item \textbf{H5-asr-dpo:} Inverted-harmlessness DPO adapters show elevated ASR versus healthy baselines ($\Delta \geq 0.10$).
    \item \textbf{H5-asr-steering:} Steering-derived adapters show elevated ASR versus healthy baselines.
    \item \textbf{H5-ordinal:} Harmful compliance increases monotonically with DPO step count within the inverted-harmlessness track.
    \item \textbf{H5-geo-behavior:} Phase-3 weight-space drift probability correlates with HEx-PHI ASR ($\rho \geq 0.60$).
\end{itemize}

\subsection{Adapter Population}

\begin{table}[H]
\centering
\caption{Adapter population.  DPO arms share all hyperparameters;
the only contrast is the HH-RLHF data axis.  Legacy adapters from a
prior POC are included but not used in primary analyses.}
\label{tab:population}
\small
\begin{tabular}{@{}llrl@{}}
\toprule
\textbf{Category} & \textbf{Method} & \textbf{$n$} &
\textbf{Intensity levels} \\
\midrule
Healthy baseline   & SFT (HH-RLHF chosen)       & 10 & Varied seeds \\
Inv.\ harmlessness & DPO (inverted harm.\ prefs) &  8 & 50--2000 steps \\
Inv.\ helpfulness  & DPO (inverted help.\ prefs) &  6 & 50--2000 steps \\
Refusal steering   & Activation $\to$ LoRA       &  6 & Varied coefficients \\
Sycophancy steering (OOD) & Activation $\to$ LoRA &  4 & Held out \\
Legacy (POC)       & DPO (gradient norm)          &  4 & Mixed \\
\midrule
\textbf{Total}     &                              & \textbf{38} & \\
\bottomrule
\end{tabular}
\end{table}

\subsection{Deviations from Pre-Registered Plan}

Centroid distance tracking produced an empty output file, so the
centroid-degradation hypothesis was not formally evaluated.  This
gap is partially covered by two results: Phase~0 showed norms are
perfectly step-monotonic and cannot carry objective identity, and
the magnitude-only feature split in Phases~1--3 reached the same
conclusion with the full population.

\section{Results}
\label{sec:results}

\subsection{Preflight: Magnitude vs.\ Shape}
\label{sec:preflight}

Before the main manufacturing run, a preflight study at $N = 6$ (4
inverted-harmlessness, 2 inverted-helpfulness adapters) tested
whether DPO objectives are geometrically separable.

Shape features (stable rank, SV entropy, effective rank) achieved
leave-one-out AUC~1.00 for objective separation.  Magnitude features
(Frobenius norm, spectral norm) achieved AUC~0.275---below chance.
Step-matched analysis confirmed the two objectives produce
identical magnitude profiles; they diverge only in shape and
direction.  At the module level, \texttt{q\_proj} alone scored
AUC~0.00 for objective separation (all signal inverted), confirming
that query-projection magnitude carries training-intensity signal,
not objective identity.

Conclusion: shape and direction features are necessary for
objective fingerprinting.  The main experiment was designed with
this constraint.

\subsection{Binary Drift Detection}
\label{sec:binary}

Healthy vs.\ all-drifted: AUC~$= 1.00$, bootstrap 95\% CI
$[1.00, 1.00]$, 23 training / 11 test adapters, zero
misclassifications.  The degenerate CI reflects perfect separation
at this sample size, not infinite precision; real uncertainty comes
from the small per-class $n$ (2--5 in the test split).

\subsection{Objective Identification}
\label{sec:objective}

All six pairwise drift-type comparisons achieve AUC~1.00
(Table~\ref{tab:pairwise}).  The hardest comparison: DPO inverted
harmlessness vs.\ DPO inverted helpfulness, same method and
hyper-parameters, only the data axis differs, also reaches 1.00.

\begin{table}[H]
\centering
\caption{Pairwise classification AUC between adapter categories.
All comparisons use logistic regression on the full feature set
with stratified 70/30 splits and bootstrap CIs.}
\label{tab:pairwise}
\small
\begin{tabular}{@{}lcc@{}}
\toprule
\textbf{Comparison} & \textbf{AUC} & \textbf{95\% CI} \\
\midrule
Healthy vs.\ Inv.\ Harmlessness (DPO) & 1.00 & [1.00, 1.00] \\
Healthy vs.\ Inv.\ Helpfulness (DPO)  & 1.00 & [1.00, 1.00] \\
Healthy vs.\ Refusal Steering         & 1.00 & [1.00, 1.00] \\
Inv.\ Harmlessness vs.\ Inv.\ Helpfulness & 1.00 & [1.00, 1.00] \\
Inv.\ Harmlessness vs.\ Refusal Steering  & 1.00 & [1.00, 1.00] \\
Inv.\ Helpfulness vs.\ Refusal Steering   & 1.00 & [1.00, 1.00] \\
\bottomrule
\end{tabular}
\end{table}

\paragraph{Severity ranking.}
Spearman $\rho$ between classifier drift scores and training step
count within each drift type: inverted harmlessness $\rho = 0.976$,
inverted helpfulness $\rho = 1.000$, refusal steering
$\rho = 0.956$ (all $p < 0.01$).

\subsection{Module Specialization}
\label{sec:module}

On binary detection (healthy vs.\ drifted), both \texttt{q\_proj}
alone and \texttt{v\_proj} alone achieve AUC~1.00: the task
saturates both modules independently.

The within-method objective split (inverted harmlessness vs.\
inverted helpfulness) is hard enough to reveal the module structure
(Table~\ref{tab:module}).

\begin{table}[t]
\centering
\caption{Module-split classification for the hardest within-method
comparison: DPO inverted harmlessness vs.\ DPO inverted
helpfulness.}
\label{tab:module}
\small
\begin{tabular}{@{}lcc@{}}
\toprule
\textbf{Module split} & \textbf{AUC} & \textbf{95\% CI} \\
\midrule
\texttt{q\_proj} only  & 0.50 & [0.00, 1.00] \\
\texttt{v\_proj} only  & 0.83 & [0.25, 1.00] \\
Both combined           & 1.00 & [1.00, 1.00] \\
\bottomrule
\end{tabular}
\end{table}

\texttt{q\_proj} carries the ``something changed'' signal;
\texttt{v\_proj} carries the ``what kind of thing changed'' signal.
This specialization was not predicted and emerges only on a task
hard enough that neither module saturates independently.

The wide CIs ($n_\text{test} = 5$) limit confidence in the exact
point estimates; the qualitative separation pattern is the finding.

\subsection{Cross-Method Generalization}
\label{sec:cross}

A binary classifier trained on DPO-drifted vs.\ healthy adapters,
tested on steering-derived adapters: AUC~$= 0.00$
($n_\text{bootstrap} = 972$, CI $[0.00, 0.00]$).  The same result
holds for the out-of-distribution steered-sycophancy adapters (same
objective as training data, different manufacturing method).

AUC~0.00 is not a null result.  It indicates perfect discriminative
power with inverted labels: every steering adapter is classified as
more healthy than every DPO adapter, with maximum confidence.  DPO
and steering perturb weight space in geometrically opposite
directions relative to the healthy centroid.

This inversion has a structural explanation. DPO produces weight deltas through iterative gradient descent, accumulating updates across many singular directions and converging toward a spectral profile similar to healthy adapters at high step counts. Activation steering is algebraic: a contrastive direction is SVD-decomposed and injected into LoRA matrices at a fixed scale, producing an effectively rank-1 perturbation that scales linearly with the steering coefficient. The classifier learned the spectral signature of gradient-based perturbations; algebraically constructed perturbations occupy the opposite region in feature space.

\textbf{Deployment implication:} A DPO trained weight monitor can only be relied upon to flag DPO trained adapters. 

\subsection{Training Objective as PC1}
\label{sec:pca_results}

PCA on $\sim$352M-dimensional flattened weight deltas of the 18 DPO
adapters:

\begin{itemize}[leftmargin=*,itemsep=2pt]
  \item \textbf{PC1} explains 20.6\% of variance and perfectly
  separates inverted harmlessness from inverted helpfulness
  (type AUC~$= 1.00$), with zero correlation to training step count
  ($\rho = -0.056$, $p = 0.83$).
  \item \textbf{PC2} captures training magnitude
  ($\rho = 0.589$, $p = 0.01$) but does not separate objectives
  (type AUC~$= 0.55$).
\end{itemize}

The training objective is the dominant axis of variation in
weight-delta space.  Two DPO populations trained with identical
hyperparameters, differing only in which preference axis is inverted,
occupy opposite poles of PC1.  The direction of the perturbation
matters more than its magnitude.

\begin{figure}[t]
\centering
\includegraphics[width=\linewidth]{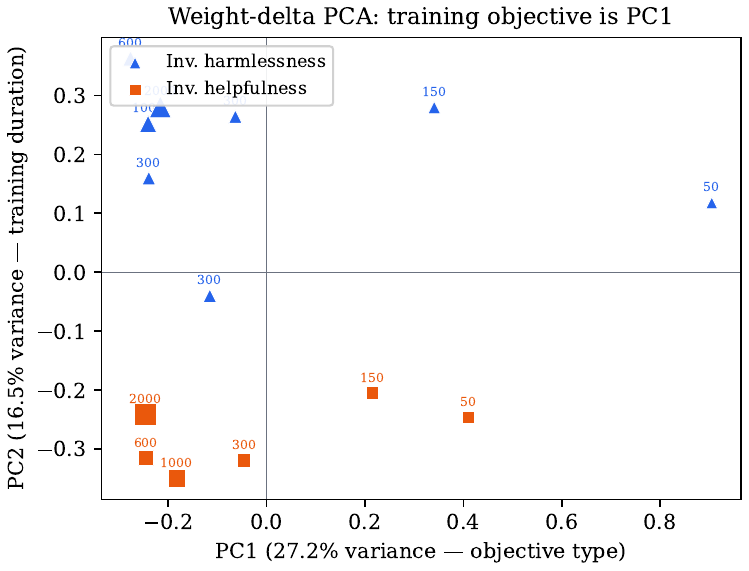}
\caption{Principal component projection of flattened DPO weight
deltas.  PC1 separates training objective (AUC~1.00); PC2 tracks
training duration ($\rho = 0.589$).  Objective identity is the
dominant axis of variation, not training intensity.}
\label{fig:pca}
\end{figure}

\subsection{Independent Convergence in Weight and Activation Space}
\label{sec:convergence}

Two structurally independent analyses both recover objective identity with perfect accuracy. Per-layer linear probes trained on hidden-state activations classify objective type at accuracy 1.00 at every layer. Weight-delta PCA, operating entirely in parameter space without inference, achieves the same separation (AUC 1.00). Neither method has access to the other's signal: the probes read activations; PCA reads weight matrices.

The two methods agree on the answer but share no geometric similarities:
\begin{equation}
  \max_{\ell, i} \left|
  \cos\!\left(n^{(\ell)}, u_i^{(\ell)}\right)
  \right| \approx 0.098
\end{equation}
across all layers, adapters, and singular-vector indices, with mean alignment ratio ~0.015.

This near-orthogonality is expected rather than surprising. Weight deltas are a static record of what gradient descent did to the parameters. Activation directions are the dynamic consequence of those changed parameters interacting with inputs at inference time. The relationship between them is mediated by the forward pass: 28 layers of matrix multiplications, nonlinearities, attention, and residual connections, which transforms the weight perturbation nonlinearly before it manifests as an activation difference. The two signals are different representations of the same external pressure (adaptation to the training objective), expressed through different stages of the model's computation.

The near-orthogonality strengthens the finding: independent geometric paths to the same classification provide stronger evidence that objective identity is a robust property of the adapted model, not an artifact of either analysis method.

The practical implication is that Weight inspection requires no inference and can be applied at the point of adapter upload; activation probing operates at inference time, providing a complementary detection window at a different stage of the model lifecycle and, given their geometric independence, are unlikely to share blind spots.

\subsection{Behavioral Evaluation (Phase 5)}
\label{sec:phase5}

\subsubsection{DPO Inverted Harmlessness}

\begin{itemize}
    \item \textbf{H5-asr-dpo: Supported.}  Mean HEx-PHI ASR = 0.266 vs.healthy mean 0.112 (elevation $+0.154$, above the pre-registered
$+0.10$ threshold).  GPT-4o calibration confirms the direction
(harm-rate elevation $+0.113$ vs.\ healthy).
    \item \textbf{H5-ordinal: Supported.}  Spearman $\rho$(step, ASR) $= 0.9856$ across six step levels, with a plateau at 1000--2000 steps
\end{itemize}

\begin{figure}[H]
\centering
\includegraphics[width=\linewidth]{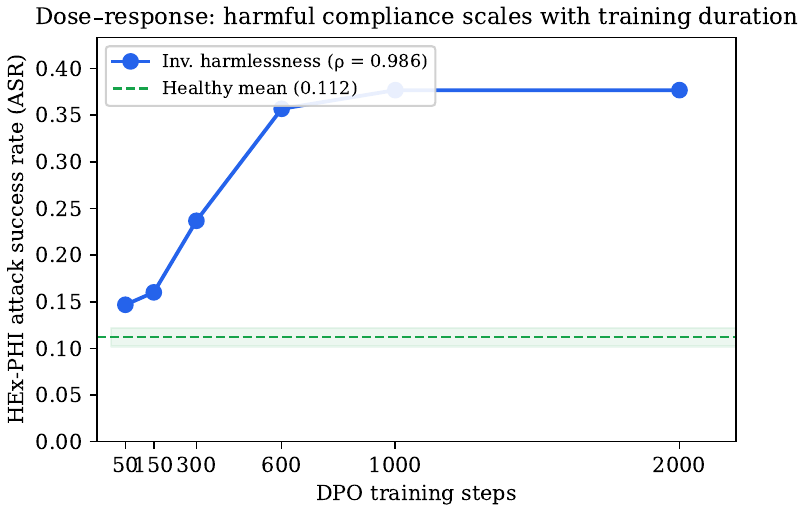}
\caption{Dose--response relationship between DPO training duration
and harmful compliance (HEx-PHI ASR) for inverted-harmlessness
adapters.  Spearman $\rho = 0.986$ across six step levels.  Healthy
baseline (dashed) at ASR = 0.112.}
\label{fig:dose}
\end{figure}

\subsubsection{DPO Inverted Helpfulness}

Mean ASR = 0.153, elevation $+0.041$, below the $+0.10$ threshold.
No dose--response ($\rho = 0.37$, $p \approx 0.47$).  GPT-4o
agrees the lift is negligible ($+0.028$).

This is consistent with a measurement mismatch: HEx-PHI evaluates
harmful compliance, not helpfulness erosion.  The geometric
classifier detects both DPO objectives as drifted, but only inverted
harmlessness produces harm on the instrument used.  The
geometry-to-behavior chain is objective-conditional: it requires
matching the evaluation suite to the manufactured objective.

\subsubsection{Steering-Derived Adapters}

\textbf{H5-asr-steering: Technically passed; substantively
invalid.}  Language generation collapsed on all steered adapters at
all intensities tested.  Llama-Guard classified degenerate token
repetition as ``unsafe,'' producing inflated ASR.  GPT-4o scored
0/300 steered responses as harmful, confirming the output is
incoherent.  The steering-to-LoRA injection method
is unsuitable for Llama-3.2-3B at the intensities tested.

\subsubsection{Geometry--Behavior Correlation}

\textbf{H5-geo-behavior: Supported.}  Spearman $\rho$ between
Phase-3 drift probability and HEx-PHI ASR = $0.72$ on 24 clean
adapters (DPO + healthy), clearing the pre-registered $0.60$
threshold ($p < 0.001$).

Including the six steered-refusal adapters inflates the correlation
to 0.84, but their ASR is a Guard artifact from generation collapse,
not a behavioral signal.

Healthy adapters cluster near drift probability 0.001; DPO adapters
near 0.999.  The rank correlation primarily reflects ``which side of
the boundary,'' not fine-grained severity within DPO.

\textbf{Exploratory:} Within inverted-harmlessness adapters, Frobenius norm of the weight delta predicts harmful compliance with near-perfect rank correlation ($\rho \approx 0.99$).  This suggests that once the drift type is known, a simple magnitude statistic can estimate behavioral severity without inference. The same relationship could not be evaluated within inverted-helpfulness adapters ($\rho \approx 0.37$), we attribute this to instrument mismatch, as the failure mode produced by inverted-helpfulness training (helpfulness erosion) is not what HEx-PHI measures (harmful compliance). The low correlation reflects an evaluation gap, not absence of a geometry–behavior link for that objective.

\begin{figure}[t]
\centering
\includegraphics[width=\linewidth]{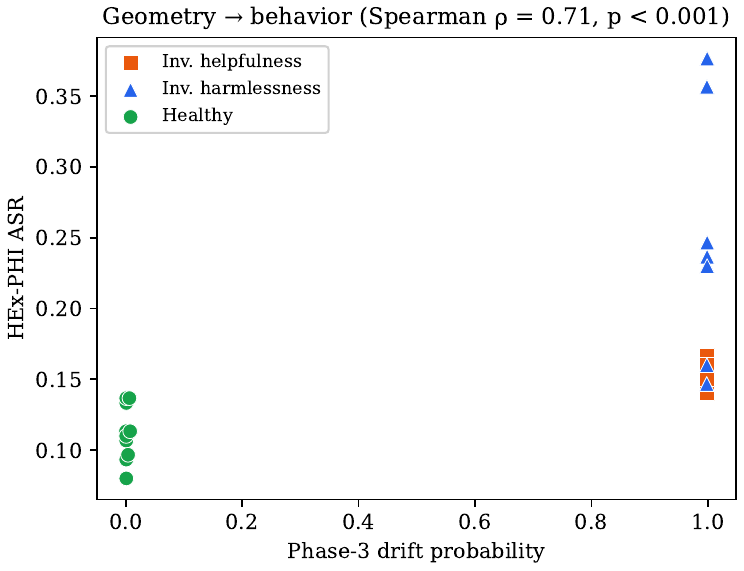}
\caption{Geometry--behavior relationship.  Phase-3 weight-space
drift probability vs.\ HEx-PHI attack success rate for 24
non-steered adapters ($\rho = 0.72$, $p < 0.001$).  The correlation
is driven by the healthy/drifted boundary; within the DPO cluster,
only inverted-harmlessness adapters show elevated ASR.}
\label{fig:geo_behavior}
\end{figure}

\subsection{Feature Importance}
\label{sec:features_importance}

The direction in which weights move away from the healthy centroid is the most informative signal for classification. In the logistic regression classifier, direction features (singular-vector cosine similarity to the healthy centroid) carry $10\times$ higher mean coefficient than shape features (stable rank, entropy, effective rank) and $30\times$ higher than magnitude features (Frobenius norm, spectral norm). This is consistent with the PCA finding: objective identity lives in the direction of the weight delta, not its magnitude or spectral shape.

\subsection{Summary of Results}

\begin{table}[H]
\centering
\caption{Summary of pre-registered and exploratory results.}
\label{tab:summary}
\small
\begin{tabular}{@{}p{5.2cm}lp{3.5cm}@{}}
\toprule
\textbf{Finding} & \textbf{Metric} & \textbf{Notes} \\
\midrule
Binary detection & AUC 1.00 & CI [1.00, 1.00] \\
All pairwise objectives & AUC 1.00 & 6/6 comparisons \\
Ordinal severity (inv.\ harm.) & $\rho$ = 0.976 & $p < 0.01$ \\
Ordinal severity (inv.\ help.) & $\rho$ = 1.000 & $p < 0.01$ \\
Ordinal severity (steering) & $\rho$ = 0.956 & $p < 0.01$ \\
Cross-method generalization & AUC 0.00 & Systematic inversion \\
Objective type = PC1 & AUC 1.00 & 20.6\% variance \\
\texttt{q\_proj} objective split & AUC 0.50 & At chance \\
\texttt{v\_proj} objective split & AUC 0.83 & CI [0.25, 1.00] \\
Weight--activation alignment & cos $\sim$0.098 & Max across all layers \\
H5-asr-dpo (ASR elevation) & $+$0.154 & Threshold: $+$0.10 \\
H5-ordinal (dose--response) & $\rho$ = 0.986 & 6 step levels \\
H5-geo-behavior & $\rho$ = 0.72 & $N=24$, $p < 0.001$ \\
Frob.\ vs.\ ASR (inv.\ harm.) & $\rho \approx 0.99$ & Exploratory \\
\bottomrule
\end{tabular}
\end{table}

\section{Discussion}
\label{sec:discussion}

\subsection{Objective Fingerprinting within a Regime}

Within a controlled DPO manufacturing regime, LoRA weight-space geometry is highly informative (Section~\ref{sec:objective}): objective type, intensity ordering,
and module-level decomposition of detection vs.\ diagnosis are all recoverable from spectral features. Direction features dominate; magnitude features track training
intensity but are blind to objective identity.

PCA on the full weight deltas (Section~\ref{sec:pca_results}) produced the stongest result. PC1, the leading axis of variation was the training objective (AUC~1.00), with no correlation to training duration ($\rho = -0.056$). Training duration loaded onto PC2. Two populations sharing identical hyperparameters, optimizer, learning rate, and LoRA configuration, differing only in which preference axis was inverted, produce weight perturbations that diverge primarily along the objective axis. This was pre-registered as 'uncertain, leaning fail', the predicted ordering was intensity first, objective on a later component.

\subsection{Cross-Method Failure as a Structural Finding}

The AUC~0.00 cross-method result (Section~\ref{sec:cross}) reveals a
geometric boundary between DPO-trained and steering-derived adapters:
a DPO-trained classifier assigns every steering adapter a lower drift
score than every DPO adapter, with maximum confidence.  The two
manufacturing methods produce structurally different perturbations: gradient-accumulated and multi-directional for DPO; Algebraic and
effectively rank-1 for steering, and the classifier learned one
structure, not both.
However, the steering-derived adapters also failed to produce coherent
generation at all tested intensities (Section~\ref{sec:phase5}),
raising a confound: the geometric opposition may reflect a broken
injection method rather than a general property of non-DPO training
regimes.  Whether functioning alternative methods (e.g., KTO, ORPO,
or properly calibrated steering on a model that tolerates it) would
also produce geometric inversion is unknown.

\textbf{Implication:}  A production weight-space monitor implementation should 
consider calibration per training method.  A single-head classifier trained
on DPO signatures would miss steering-based attacks and actively classify them as safe. A viable architecture is likely a multi-head monitor: one head per known manufacturing method, plus an anomaly-detection fallback for unknown methods.

\subsection{The Geometry--Behavior Chain}

The geometry-to-behavior link at $\rho = 0.72$ is a proof of concept
for early detection (Section~\ref{sec:phase5}).  Weight-space
structure can flag which side of the drift boundary an adapter is on
before any behavioral evaluation runs.  The link is
objective-conditional: it works for inverted harmlessness (where
HEx-PHI matches the failure mode) and does not work for inverted
helpfulness (where the evaluation suite does not match).  A
practical system requires matching the behavioral instrument to the
training objective.

The Frobenius-vs.-ASR correlation within inverted-harmlessness
adapters ($\rho \approx 0.99$) suggests that fine-grained severity
estimation is possible within a single objective, but only when the
evaluation suite is correctly matched.  This motivates a two-stage
detection architecture: classify drift type first, then estimate
severity within type using type-specific magnitude features.

\subsection{Connection to Persona Selection Model}

The Persona Selection Model~\citep{marks2026psm} argues that
post-training selects persona traits from a structured space already
present in the pre-trained model.  Our results provide independent
weight-space evidence consistent with this framework:

\begin{itemize}[leftmargin=*,itemsep=2pt]
  \item Training objective (which trait is being selected) is the
  dominant geometric axis, not training intensity.
  \item \texttt{q\_proj}/\texttt{v\_proj} specialization
  (detection vs.\ diagnosis) suggests persona traits are distributed
  across functionally distinct attention components.
  \item Both weight space and activation space encode objective identity perfectly, yet share almost no geometric structure, consistent with persona-relevant information being encoded across multiple independent representational subspaces.
\end{itemize}

We do not validate PSM predictions directly.  The geometry described
here is independently consistent with structured persona space.

\subsection{GPT-4o Calibration Findings}

Three blind spots appear exclusively in the highest-step
inverted-harmlessness adapter (2000 steps): two cases of
refusal-with-loophole (the model declines the explicit request,
then offers related assistance serving the same goal) and one case
of partial compliance with actionable content.  These do not appear
in healthy adapters and may represent drift-specific evasion
patterns.

Additionally, the calibration pass revealed that Llama-3.2-3B-Instruct complies
with certain Category~3 (Hate/ Harass/ Violence) prompts when framed as
educational or analytical: a base-model behavior present in healthy
adapters, not introduced by fine-tuning.  All 17 Guard blind spots in
this category trace to two such prompts.  Llama-Guard-3-1B does not
flag these consistently, requiring GPT-4o calibration to distinguish
base-model compliance from drift-induced harm.

\section{Limitations}
\label{sec:limitations}

\begin{enumerate}[leftmargin=*,itemsep=3pt]

  \item \textbf{Single base model.}  All results are on
  Llama-3.2-3B-Instruct.  Generalization to other architectures,
  model sizes, or full fine-tuning (rather than LoRA rank~8) is
  unknown.

  \item \textbf{Small per-class $n$.}  Test-set sizes of 2--5
  per class produce degenerate bootstrap CIs at perfect
  classification.  The current tasks may be too easy; the
  interesting experiment is the one where AUC drops.

  \item \textbf{Manufactured drift only.}  All adapters are
  deliberately constructed.  Generalization to organic drift from
  distributional shift, data contamination, or mixed-objective
  training is untested and likely harder.

  \item \textbf{Guard model.}  Behavioral scoring used
  Llama-Guard-3-1B instead of the pre-registered Guard-3-8B.
  GPT-4o calibration was required to avoid a false
  positive---Guard-1B classified steered-adapter incoherent output as
  ``unsafe.''  The geometry--behavior correlation excludes steered
  adapters for this reason.

  \item \textbf{Behavioral suite mismatch.}  HEx-PHI is the wrong
  instrument for inverted-helpfulness adapters.  The null behavioral
  result on that objective may reflect instrument mismatch, not
  absence of behavioral drift.

  \item \textbf{Centroid distance.}  The pre-registered
  centroid-degradation hypothesis was not evaluated due to a
  pipeline error.

  \item \textbf{Superposition.}  Mixed-objective adapters (e.g.,
  simultaneous inverted harmlessness and helpfulness) are not tested.
  Whether geometric signatures decompose linearly is unknown.
\end{enumerate}

\section{Future Work}
\label{sec:future}

\paragraph{Early detection.}  Does the geometry signal appear
before behavior diverges?  Testing whether spectral signatures are
present at step counts where behavioral evaluation cannot yet
distinguish drifted from healthy adapters.

\paragraph{Severity estimation.}  The geometry--behavior link at
$\rho = 0.72$ classifies the boundary, not the severity.  A
practical early-warning system needs within-type severity ranking
with matched evaluation suites.

\paragraph{Matrix coverage.}  This experiment covers only
\texttt{q\_proj} and \texttt{v\_proj}.  Abliteration-style attacks
target \texttt{o\_proj} and \texttt{down\_proj}.  Expanding matrix
coverage or using a multi-head detector per matrix family is a
natural extension.

\paragraph{Cross-method features.}  The cross-method failure
motivates representation-space features as a method-agnostic
alternative.  This is downstream of disambiguating whether the
steering generation collapse is a property of the injection method
or of representational entanglement.

\paragraph{Second base model.}  Architectural generalization is the
largest unknown.  The next model should be architecturally
different, not merely a larger Llama variant.

\paragraph{Sycophancy manufacturing.}  Building a measurably
sycophantic adapter requires preference data that cleanly labels
agreeableness vs.\ honesty.  No public dataset currently provides
this; purpose-built synthetic preference data is needed.

\section{Conclusion}
\label{sec:conclusion}

We conducted a pre-registered experiment on 38 LoRA adapters across
four manufacturing categories on Llama-3.2-3B-Instruct.  The
results establish three findings:

\begin{enumerate}[leftmargin=*,itemsep=2pt]
  \item Within a controlled DPO regime, the spectral geometry of
LoRA weight deltas encodes training objective.  A logistic
classifier perfectly separates all pairwise objectives (AUC~1.00), and training objective is
the dominant axis of variation in weight-delta space (PC1), orthogonal to training duration.

  \item The geometry-to-behavior link is objective-specific: for inverted-harmlessness training, weight-space drift probability correlates with harmful compliance at
  $\rho = 0.72$ with near-perfect dose--response ($\rho = 0.986$).

  \item Both weight space and activation space encode objective identity, yet they share almost no geometric structure (cosine $\sim$0.098). Objective information is distributed and representation-dependent: weight-space and activation-space monitors would operate on independent signals.
\end{enumerate}

Weight-space geometry carries more fine-tuning provenance than summary statistics suggest, enough to identify what a model was trained to do, and to flag when that training may cause harm.

\section*{Acknowledgments}

This work was conducted independently with personal funding.
Development used Cursor as the primary IDE and Claude (Anthropic)
for research iteration and code development. We thank the developers of the open-source
tools used in this work: Hugging Face Transformers, PEFT, TRL, and the Llama model family.

\bibliographystyle{plainnat}

\appendix

\section{Proof-of-Concept Details}
\label{app:poc}

The proof-of-concept (POC) used seven adapters: three healthy SFT
baselines and four DPO-drifted adapters trained for 50, 150, 300,
and 600 steps on inverted harmlessness preferences from HH-RLHF.
A logistic regression classifier on per-layer spectral features
achieved AUC~1.00 with leave-one-out cross-validation. A
15-scenario behavioral test suite targeting sycophancy scored
AUC~0.083 on the same population: below chance.

The POC revealed the convergence problem: at 600 DPO steps, all
summary statistics (Frobenius norm, effective rank, stable rank)
converge to within 4\% of the healthy mean, while behavioral
divergence is maximal.  The classifier still succeeds because the
discriminative signal lives in per-layer patterns, not summary
aggregates.

This observation motivated the transition from anomaly detection
(``does this adapter look abnormal?'') to objective fingerprinting
(``what was this adapter trained to do?'').

\section{Phase 0: Signal Localization}
\label{app:phase0}

Using the spectral feature matrix from the POC, we tested where
signal lives by depth, by module, and along the DPO step ramp
(50--600 steps).

\textbf{Depth:}  All layer depths contributed to drift separation
uniformly.  Drift signal is not concentrated in early, middle, or
late layers.

\textbf{Module:}  \texttt{q\_proj} is reliable for healthy-vs.-DPO
separation.  \texttt{v\_proj} performed at chance during Phase~0.

\textbf{Magnitude monotonicity:}  Frobenius and spectral norms
track DPO step count monotonically at almost every sublayer
($\rho \approx +1.0$).  Magnitude features encode ``how much
training happened,'' not ``what kind of training.''

\section{Supplementary Figures}
\label{app:figures}

Figures~\ref{fig:pca}--\ref{fig:geo_behavior} appear in the main
text.  The following supplementary figures provide an additional
visualization of the results.

\begin{figure}[H]
\centering
\includegraphics[width=0.7\linewidth]{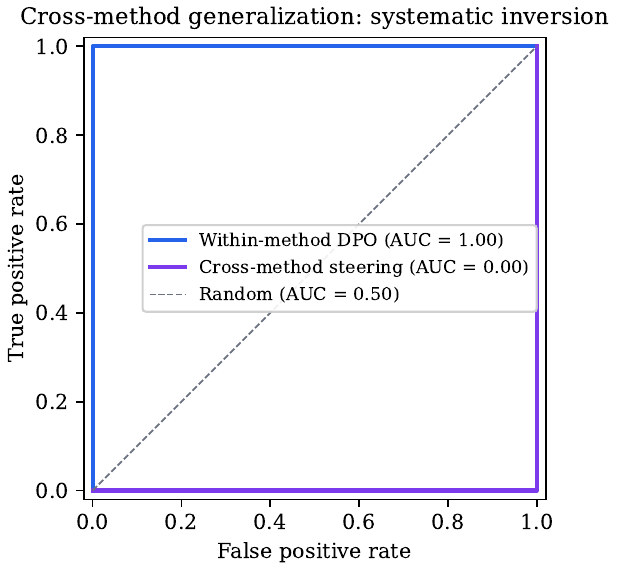}
\caption{Cross-method generalization.  Within-method DPO classifier
achieves AUC~1.00 (perfect ROC).  The same classifier tested on
steering-manufactured adapters produces AUC~0.00 (systematic
inversion, not noise).}
\label{fig:cross_roc}
\end{figure}

\begin{figure}[H]
\centering
\includegraphics[width=0.85\linewidth]{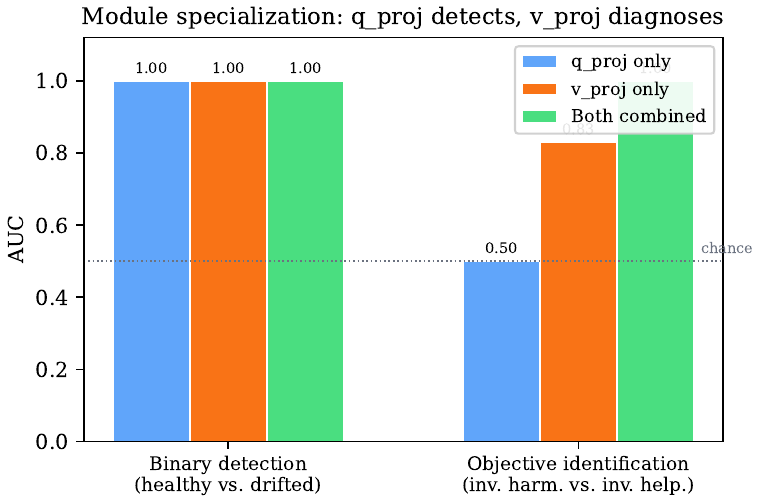}
\caption{Module specialization across task difficulty.  On binary
detection (left), both modules saturate independently.  On the
harder objective-identification task (right), \texttt{q\_proj}
drops to chance while \texttt{v\_proj} carries the diagnostic
signal.}
\label{fig:module_split}
\end{figure}

\begin{figure}[H]
\centering
\includegraphics[width=0.75\linewidth]{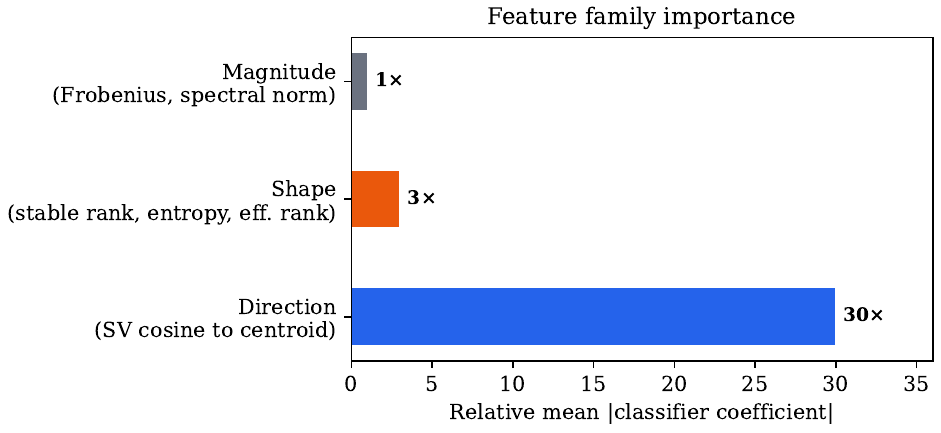}
\caption{Relative importance of feature families.  Direction
features (singular-vector cosine to healthy centroid) dominate
by $30\times$ over magnitude features and $10\times$ over shape
features.}
\label{fig:feat_importance}
\end{figure}

\begin{figure}[H]
\centering
\includegraphics[width=\linewidth]{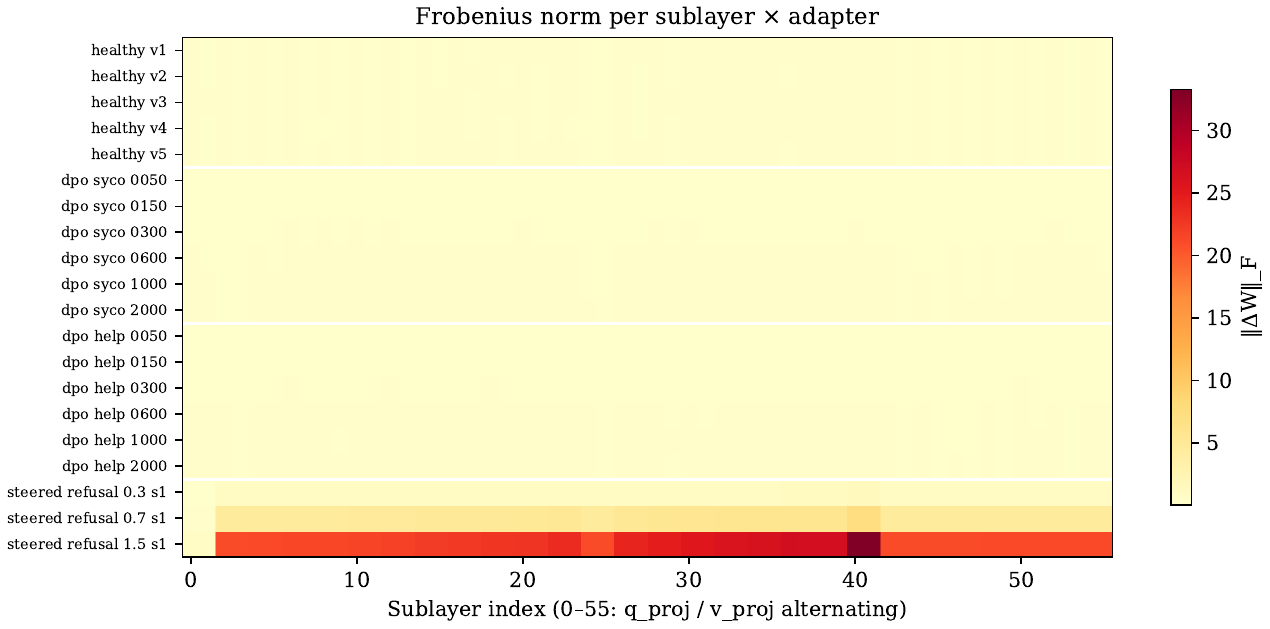}
\caption{Frobenius norm per sublayer across adapters, ordered by
category and training intensity.  Magnitude increases monotonically
with step count within each type but does not distinguish between
objectives---consistent with the finding that magnitude features
cannot carry objective identity.}
\label{fig:heatmap}
\end{figure}

\begin{figure}[H]
\centering
\includegraphics[width=0.85\linewidth]{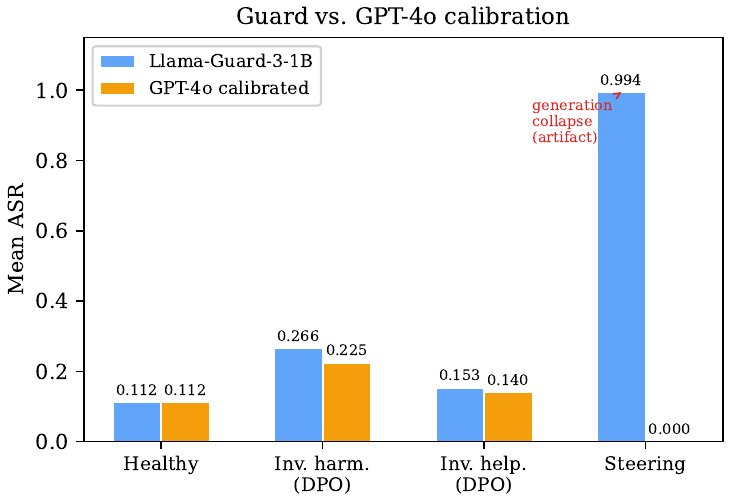}
\caption{Llama-Guard-3-1B vs.\ GPT-4o calibrated ASR by adapter
category.  Guard and GPT-4o agree on healthy and DPO adapters.
For steering-derived adapters, Guard classifies degenerate output
as ``unsafe'' (ASR~0.994) while GPT-4o correctly identifies
generation collapse (ASR~0.0).}
\label{fig:calibration}
\end{figure}

\section{Data and Code Availability}
\label{app:data}

Pre-registration documents, exit reports, spectral feature matrices,
classification results, and harmfulness evaluation summaries are
available at:
\begin{center}
\url{https://github.com/roip/task-geometry-experiment-results}
\end{center}
LoRA adapter weights are withheld because the drifted adapters produce
measurably more harmful outputs; released spectral features and
ASR summaries are sufficient to verify all statistical claims.
The analysis code is available upon request.

\end{document}